\documentclass[sigconf]{acmart}
\AtBeginDocument{%
  }

\setcopyright{acmlicensed}
\copyrightyear{2018}
\acmYear{2018}
\acmDOI{XXXXXXX.XXXXXXX}
\acmConference[Conference acronym 'XX]{Make sure to enter the correct
  conference title from your rights confirmation email}{June 03--05,
  2018}{Woodstock, NY}
\acmISBN{978-1-4503-XXXX-X/2018/06}

\usepackage{multirow}
\usepackage[table]{xcolor}
\usepackage{colortbl}
\usepackage{enumitem}
\usepackage[most]{tcolorbox}
\usepackage{hyperref}

\begin{document}

\title{LawThinker: A Deep Research Legal Agent in Dynamic Environments}

\author{Xinyu Yang}
\email{yxygsai@ruc.edu.cn}
\affiliation{%
  \institution{Renmin University of China}
  \city{Beijing}
  \country{China}
}

\author{Chenlong Deng}
\email{dengchenlong@ruc.edu.cn}
\affiliation{%
  \institution{Renmin University of China}
  \city{Beijing}
  \country{China}
}

\author{Tongyu Wen}
\email{wentongyu@ruc.edu.cn}
\affiliation{%
  \institution{Renmin University of China}
  \city{Beijing}
  \country{China}
}

\author{Binyu Xie}
\email{xiebinyu0929@ruc.edu.cn}
\affiliation{%
  \institution{Renmin University of China}
  \city{Beijing}
  \country{China}
}

\author{Zhicheng Dou}
\authornote{Zhicheng Dou is the corresponding author.}
\email{dou@ruc.edu.cn}
\affiliation{%
  \institution{Renmin University of China}
  \city{Beijing}
  \country{China}
}
% \thanks{Corresponding author.}

\renewcommand{\shortauthors}{Trovato et al.}

\begin{abstract}
Legal reasoning requires not only correct outcomes but also procedurally compliant reasoning processes. However, existing methods lack mechanisms to verify intermediate reasoning steps, allowing errors such as inapplicable statute citations to propagate undetected through the reasoning chain. To address this, we propose \textbf{LawThinker}, an autonomous legal research agent that adopts an Explore-Verify-Memorize strategy for dynamic judicial environments. The core idea is to enforce verification as an atomic operation after every knowledge exploration step. A DeepVerifier module examines each retrieval result along three dimensions of knowledge accuracy, fact-law relevance, and procedural compliance, with a memory module for cross-round knowledge reuse in long-horizon tasks. Experiments on the dynamic benchmark J1-EVAL show that LawThinker achieves a 24\% improvement over direct reasoning and an 11\% gain over workflow-based methods, with particularly strong improvements on process-oriented metrics. Evaluations on three static benchmarks further confirm its generalization capability. The code is available at \url{https://github.com/yxy-919/LawThinker-agent}.
% https://anonymous.4open.science/r/LawThinker-9C24
% Recent advances in Large Reasoning Models (LRMs) have shown impressive multi-step problem-solving capabilities. However, legal reasoning poses a unique challenge: outcomes must be supported by an accurate and procedurally compliant reasoning process because intermediate errors can propagate to undermine conclusions. To address this challenge, we propose \textbf{LawThinker}, an autonomous legal research agent that integrates iterative exploration with explicit verification throughout the reasoning process. LawThinker employs 15 specialized tools across three dimensions of exploration, verification, and memorization --- to retrieve relevant legal knowledge, check the accuracy, relevance, and procedural correctness of intermediate reasoning, and store key information for long-horizon tasks. In particular, a DeepVerifier module is designed to enable the model to autonomously examine each reasoning step, preventing the accumulation of errors. Extensive experiments on the dynamic legal benchmark J1-EVAL demonstrate that LawThinker significantly outperforms existing methods and achieves both outcome accuracy and procedural consistency. Further evaluations on three static benchmarks, LawBench, LexEval, and UniLaw-R1-Eval, confirm its generalization capability. The code is available at \url{https://anonymous.4open.science/r/LawThinker-9C24}.
% Our framework enables fully process-aware legal intelligence, bridging knowledge retrieval, logical verification, and normative generation.
\end{abstract}




\maketitle

\section{Introduction}
\begin{figure}[t]
  \centering
  \includegraphics[width=\linewidth]{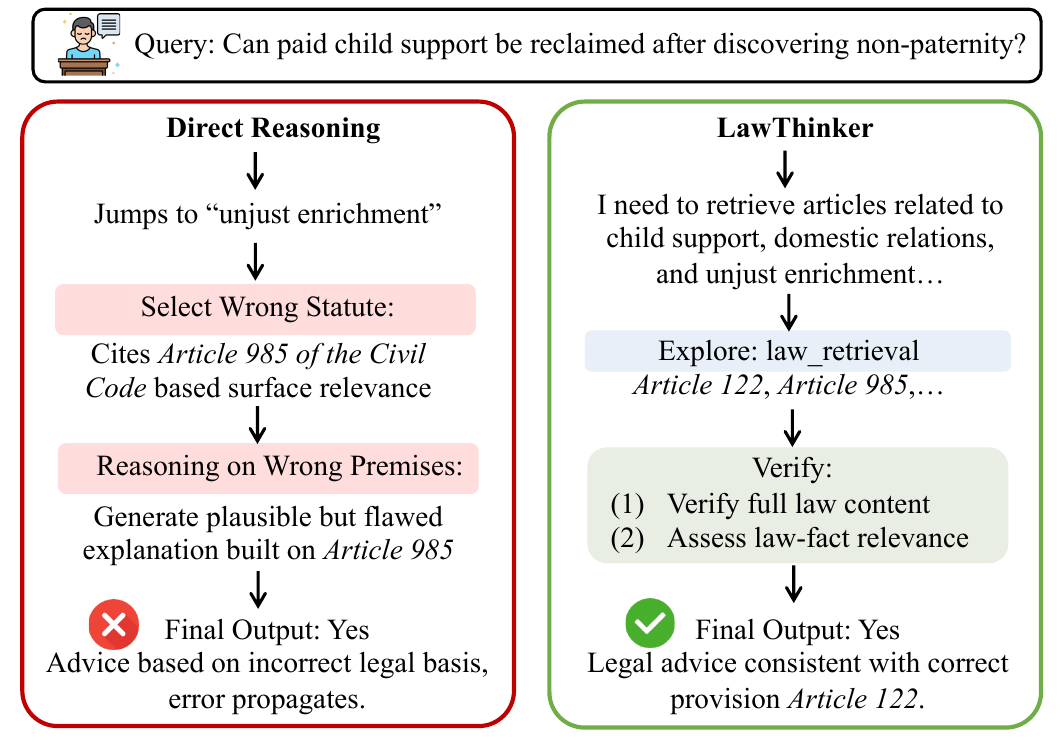}
  \caption{An example of error propagation from an incorrect statute citation.
While both methods reach the same affirmative outcome, direct reasoning cites an inapplicable article, while LawThinker mitigates this issue by exploring with explicit verification, underscoring the necessity of process-level compliance in legal reasoning beyond answer correctness.}
  \label{fig:case}
  \Description{}
\end{figure}

% Recent advances in Large Reasoning Models (LRMs) have demonstrated remarkable capabilities across a wide range of domains, including mathematics, code, and science.
Recent advances in Large Reasoning Models (LRMs) have demonstrated strong multi-step problem-solving capabilities in domains such as mathematics~\cite{wang2025survey,ahn2024large,xu2025towards,team2024qwq}, code generation~\cite{halim2025study,ding2024reasoning,jiang2024survey,jiang2024self,sun2024enhancing}, and scientific reasoning~\cite{fan2025megascience,yan2025position}. These successes have prompted growing interest in applying LRMs to legal reasoning tasks. However, legal reasoning differs from other domains in a critical way: \textit{a legally valid conclusion must be supported not only by a correct outcome, but also by a reasoning process that is accurate and procedurally compliant.} Citing an inapplicable statute or omitting a required reasoning step renders the entire analysis legally defective, even if the final answer happens to be correct. The challenge is further amplified in real-world judicial scenarios, which are inherently dynamic and interactive. Legal consultations involve multi-turn dialogues where users progressively raise follow-up questions, document drafting requires iterative information gathering across multiple exchanges, and courtroom proceedings demand strict adherence to multi-stage procedural workflows. These dynamic settings impose strong demands on the agent's ability to maintain accurate, procedurally grounded reasoning over long interaction horizons.

% By leveraging step-by-step reasoning and long-context modeling, these models are able to solve complex problems that require multi-stage logical decomposition. Such successes naturally raise expectations for their applicability to legal reasoning tasks.

% 过程很重要
% However, legal reasoning presents fundamentally different challenges. Unlike solving mathematical or coding problems, legal scenarios demand extreme precision and strict adherence to procedural logic. \textit{A legally valid conclusion is determined not only by the correctness of its outcome, but also by the legitimacy of the reasoning process from which it is derived.} This process-oriented nature distinguishes legal reasoning from many other domains where intermediate errors can be corrected locally or even tolerated without invalidating the final result.

Figure~\ref{fig:case} illustrates this challenge with a concrete example. When asked whether paid child support can be reclaimed after discovering non-paternity, a direct reasoning approach selects Article 985 of the Civil Code based on surface-level keyword matching with ``unjust enrichment'', and then constructs an apparently coherent explanation on this incorrect legal basis. Although the final answer happens to be affirmative, the cited provision is inapplicable, and the correct legal ground should be Article 122. This case reveals that errors introduced at the statute selection stage are not corrected by current legal reasoning systems, but instead absorbed into subsequent reasoning steps. As a result, the final conclusion may appear well-reasoned yet lack a valid legal foundation. More critically, such errors are difficult to detect through outcome evaluation alone, since the final answer may coincidentally be correct. This observation highlights the need for explicit verification mechanisms that can examine the accuracy and relevance of each intermediate step, rather than relying solely on end-to-end outcome correctness.

% 推理过程中的幻觉
% One critical challenge arises from hallucinations and reasoning inaccuracies in intermediate steps. Legal reasoning is often long-horizon and multi-stage, involving fact interpretation, legal element identification, and statute matching. Errors occurring at any intermediate step are prone to propagate and accumulate throughout the reasoning chain, ultimately leading to incorrect or unreliable conclusions as shown in Figure~\ref{fig:case}. This issue is particularly severe in long-horizon legal reasoning scenarios, such as courtroom simulations, case adjudication, and multi-round legal consultations. As a result, seemingly coherent reasoning chains may ultimately lead to conclusions that are plausible yet legally unsound.

Beyond error propagation, legal reasoning imposes a stricter requirement that is often overlooked by existing systems. In judicial practice, every conclusion must be supported by accurate statutory provisions and grounded in verified case facts, and the reasoning process must follow legally prescribed procedures. A judgment built on an inapplicable statute may be overturned on appeal, regardless of whether its outcome is correct. This demand for process-level legitimacy is precisely where current methods fall short. Existing approaches either rely solely on parametric knowledge with frequent hallucinations~\cite{dahl2024large,blair2025llms,yu2025evaluating}, or incorporate external retrieval without verifying the accuracy and relevance of the retrieved information~\cite{wu2023precedent,peng2024athena,zhou2026lras}. Even methods that introduce step-level verification focus on whether intermediate steps lead to a correct outcome, rather than whether the reasoning process conforms to legally prescribed procedures. These existing methods fail to jointly ensure knowledge accuracy, fact-law relevance, and procedural compliance throughout the reasoning trajectory.

% 推理过程要遵守程序正义
% Beyond correctness, legal reasoning is further constrained by the principle of procedural justice. Legal decision-making requires that the reasoning process follow prescribed procedures, identify and apply appropriate legal grounds, and remain transparent. Any procedural deviation, such as omitting essential reasoning steps, invoking inapplicable legal provisions, or drawing conclusions without sufficient justification, renders the reasoning process legally defective, even if the final outcome appears correct. Therefore, ensuring procedural compliance throughout the reasoning trajectory is therefore a fundamental requirement for trustworthy legal AI systems.

% Recent agentic approaches attempt to address these challenges by enabling models to autonomously retrieve external knowledge and interact with tools during the reasoning process. By reducing reliance on purely parametric knowledge, these methods improve factual grounding and robustness in complex tasks. However, most existing agentic frameworks primarily emphasize the correctness of final answers, while offering limited mechanisms for assessing and controlling the quality of intermediate reasoning steps. Whether intermediate conclusions are logically sound or procedurally compliant is often left unchecked, leaving the reasoning trajectory vulnerable to error propagation and procedural violations.

To address these limitations, we propose \textbf{LawThinker}, an autonomous legal research agent designed for dynamic judicial environments. LawThinker adopts an \textbf{Explore-Verify-Memorize} strategy that integrates iterative knowledge exploration with explicit verification throughout the reasoning process. When encountering knowledge gaps, the agent autonomously retrieves relevant legal provisions, cases, and procedural rules from external knowledge bases. Each retrieval is immediately followed by a \textbf{DeepVerifier} module, which examines the accuracy of the retrieved content, its relevance to the case facts, and whether its use conforms to legal procedures. The structured verification results are fed back into the reasoning chain, allowing the agent to correct errors before they propagate to subsequent steps. To support long-horizon interactive tasks, a memory module persistently stores validated legal knowledge and key case context, enabling the agent to reuse verified information across multiple reasoning turns. We further design a suite of 15 legal tools spanning the three dimensions of exploration, verification, and memorization, providing the agent with comprehensive capabilities tailored for real-world judicial scenarios.

% To bridge this gap, we propose \textbf{LawThinker}, an autonomous deep research legal agent that integrates iterative exploration with explicit verification throughout the reasoning process. LawThinker is designed to autonomously invoke \textbf{exploration} tools to acquire comprehensive legal knowledge relevant to the given case. More importantly, to equip LawThinker with deep \textbf{verification} capabilities, we introduce a \textbf{Deep Checker} module, which enables LRMs to autonomously call checking tools to examine each exploration and reasoning step. The module returns structured checking results, allowing the model to identify inaccuracies, irrelevancies, and procedural incompleteness before proceeding to subsequent steps. By doing so, it aims to produce legal conclusions that are not only accurate in outcome but also legitimate in process. In addition, LawThinker interacts with a \textbf{memory} module that supports the \textbf{storage} and fetch of key legal knowledge and case-specific contextual information. This design enables robust long-horizon reasoning, allowing the agent to repeatedly utilize verified information while avoiding redundant exploration.

We conduct extensive experiments on both dynamic and static legal benchmarks to validate the effectiveness of LawThinker. On J1-EVAL, a dynamic benchmark covering six judicial scenario types, LawThinker achieves a 24\% overall improvement over direct reasoning baselines and an 11\% gain over workflow-based methods. The improvement is particularly significant on process-oriented metrics such as format-following and procedural-following scores, confirming that step-level verification directly strengthens procedural compliance. Our analysis also reveals that workflow-based methods, despite accessing external knowledge, can exhibit lower process-oriented scores than direct reasoning, indicating that unverified retrieval may actively harm procedural correctness. In courtroom simulation scenarios, LawThinker achieves the highest stage completion rates across all civil and criminal trial phases, demonstrating its ability to maintain procedurally grounded reasoning over long interaction horizons. Further evaluations on three static benchmarks, LawBench, LexEval, and UniLaw-R1-Eval, show an average accuracy improvement of approximately 6\% over direct reasoning, confirming that the Explore-Verify-Memorize strategy generalizes beyond dynamic settings to conventional legal reasoning tasks.

In summary, our contributions are as follows:

(1) We propose LawThinker, an autonomous legal research agent that adopts an Explore-Verify-Memorize strategy for dynamic judicial environments. In particular, we design a DeepVerifier module that verifies each exploration step along three dimensions of knowledge accuracy, fact-law relevance, and procedural compliance, preventing error accumulation throughout reasoning.

(2) We design 15 legal tools spanning exploration, verification, and memorization, enabling the agent to navigate the legal knowledge space, validate intermediate reasoning steps, and reuse verified information across long-horizon tasks.

(3) Extensive experiments on a dynamic benchmark covering six judicial scenarios and three static benchmarks demonstrate that LawThinker significantly outperforms existing methods on both outcome and process-oriented metrics, with particularly strong gains in procedural compliance and courtroom stage completion.

\begin{figure*}[t]
  \centering
  \includegraphics[width=\linewidth]{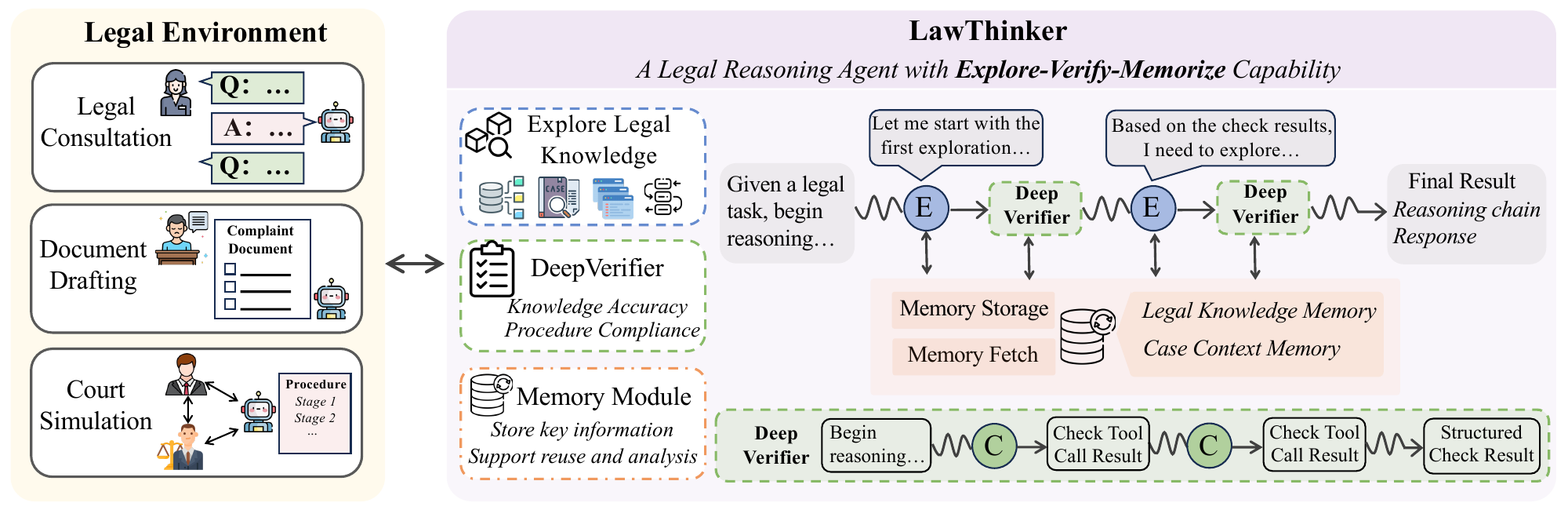}
  \caption{Overview of our autonomous legal research agent LawThinker, which adopts an Explore-Verify-Memorize strategy, integrating iterative exploration with explicit verification during reasoning and interacting closely with a memory module.}
  \label{fig:pipeline}
  \Description{}
\end{figure*}

\section{Related Work}
\subsection{LLM Reasoning and Verification}
Recent advances in Large Language Models (LLMs) have substantially enhanced reasoning capabilities in the legal domain~\cite{zhou2024survey,chen2025towards,li2025system,liu2025efficient}. Beyond general chain-of-thought (CoT)~\cite{CoT}, several studies~\cite{LoT,syler,deng2023syllogistic,wroblewski1974legal} have introduced legal syllogism reasoning to improve performance on legal tasks. Specialized frameworks have also been proposed to address the inherent complexity of judicial practice~\cite{wang2024legalreasoner,zhang2024hd,wei2025llms,yuan2026multi,yuan2024can}. For instance, ADAPT~\cite{adapt} introduces an Ask-Discriminate-Predict reasoning framework inspired by human judicial decision-making. Most recently, legal reasoning research has shifted toward agentic paradigms~\cite{wang2025mars,zhou2026lras,underkuffler1998agentic}. GLARE~\cite{glare} proposes an agentic legal reasoning framework that dynamically interacts with external modules to retrieve and apply legal knowledge during decision-making. This transition toward agentic reasoning has emerged as a leading paradigm in recent research, where models no longer reason over static prompts alone, but actively navigate the legal knowledge space to reach final judgments~\cite{wei2026agentic,o2024agentic}. 

Despite these advances, ensuring the accuracy and procedural compliance of the reasoning process itself remains a significant challenge. LegalReasoner~\cite{legalreasoner} introduces step-wise reasoning for legal judgment prediction by identifying dispute points and employs a process verifier to validate each reasoning step. However, the verification primarily focuses on whether intermediate steps contribute to a correct final conclusion, rather than whether the reasoning process adheres to legally prescribed procedures. Recent benchmarks~\cite{legalagentbench,jing2025maslegalbench,j1} reveal that evaluating only final answers is insufficient, and that the quality and validity of intermediate reasoning steps are critical to reliable legal decision-making.

% \begin{figure*}[t]
%   \centering
%   \includegraphics[width=\linewidth]{pic/pipeline6.pdf}
%   \caption{Overview of our autonomous legal research agent LawThinker, which adopts an Explore-Verify-Memorize strategy, integrating iterative exploration with explicit verification during reasoning and interacting closely with a memory module.}
%   \label{fig:pipeline}
%   \Description{}
% \end{figure*}

\subsection{LLM-based Legal Agents}
LLM-based legal applications are rapidly evolving from static, single-turn tasks to dynamic, interactive judicial scenarios~\cite{haase2025beyond,zhangsim,chen2025simulating}. SimCourt~\cite{zhang2025chinese} focuses on simulating courtroom procedures, covering the full judicial workflow from case presentation to deliberation. AgentsBench~\cite{jiang2025agentsbench} further models the discussion process of the Chinese judicial bench by leveraging multiple LLM-driven agents, simulating interactions among professional judges and lay judges. ChatLaw~\cite{cui2023chatlaw} adopts a multi-agent collaborative framework to simulate the division of labor within real-world law firms. By assigning distinct roles to multiple agents in different legal scenarios, these multi-agent systems enable role-aware reasoning that better reflects real judicial and legal practice~\cite{fang2025comprehensive,zhu2008role}. However, most existing approaches are designed for specific tasks or fixed scenarios, limiting their adaptability to diverse legal contexts. More importantly, these systems often focus on producing plausible outcomes or dialogues, while leaving intermediate reasoning steps largely unverified, thereby increasing the risk of accumulated logical errors or procedural violations in complex legal reasoning tasks. In contrast, our framework aims to support diverse legal scenarios, including legal reasoning, document drafting, and interactive courtroom simulation, while explicitly emphasizing the verification and procedural compliance of intermediate reasoning steps.

% \begin{figure*}[t]
%   \centering
%   \includegraphics[width=\linewidth]{pic/pipeline6.pdf}
%   \caption{Overview of our autonomous legal research agent LawThinker, which adopts an Explore-Verify-Memorize strategy, integrating iterative exploration with explicit verification during reasoning and interacting closely with a memory module.}
%   \label{fig:pipeline}
%   \Description{}
% \end{figure*}

\section{Methodology}
\subsection{Task Formulation}
We consider interactive legal tasks situated in dynamic judicial environments. A legal agent assumes a designated judicial role, such as a legal trainee, lawyer, or judge, and engages in multi-round dialogues with other participants in the environment. At each dialogue round $t$, given the dialogue history $H_t$ and a set of available tools $\mathcal{T}$, the agent is required to generate a reasoning chain $R_t$ along with a corresponding response $r_t$. The interaction continues until the task objective is fulfilled, which may involve resolving a legal inquiry, producing a formal legal document, or completing a judicial procedure.

% Unlike standard reasoning tasks where only the final answer is evaluated, legal tasks impose a dual requirement on the agent's output. The response $r_t$ must be outcome-correct, and the reasoning chain $R_t$ must be process-compliant, meaning that every intermediate step, including statute citations, fact-law mappings, and procedural sequences, must be legally valid. Moreover, the dynamic nature of these tasks introduces a progressive information acquisition challenge. 

The agent does not receive all relevant information at once. Instead, it must incrementally gather knowledge and context across dialogue rounds, and the reasoning at round $t$ depends on information accumulated from all previous rounds. This requires the agent to maintain accurate and verified knowledge over long interaction horizons, as errors introduced in early rounds can propagate and compromise the entire subsequent reasoning trajectory. 
% To meet this requirement, we decompose the reasoning chain $R_t$ into a sequence of steps $\{s_1, s_2, ...,s_n\}$. At each step $s_i$, the agent does not receive all relevant information at once. Instead, it must incrementally gather knowledge and context across dialogue rounds, and the reasoning at round $t$ depends on information accumulated from all previous rounds. This requires the agent to maintain accurate and verified knowledge over long interaction horizons, as errors introduced in early rounds can propagate and compromise the entire subsequent reasoning trajectory. 

To meet this requirement, we decompose the reasoning chain $R_t$ into a sequence of steps $\{s_1, s_2, ...,s_n\}$. At each step $s_i$, the agent may invoke an exploration tool and obtain a result $e_i$. A verification module then takes $(s_i, e_i, H_t)$ as input and produces a structured assessment $v_i = (a_i, \text{rel}_i, p_i)$, representing knowledge accuracy, fact-law relevance, and procedural compliance respectively. Based on $v_i$, the agent decides whether to accept the current result, revise its reasoning, or reformulate the query and re-explore. This step-level verification mechanism is the foundation of our framework, and we describe its design in detail in the following sections.

% Legal reasoning differs from mathematical or code reasoning in a fundamental way. In mathematics, a correct final answer usually suffices to validate the solution. In legal practice, however, the reasoning process itself must be accurate and procedurally compliant. A conclusion built on an inapplicable statute or derived through an incorrect procedural sequence is legal defective, even if the final outcome happens to be correct. This requirement ...(waiting..)

% \subsection{Preliminaries}
% We first formally define our tasks. Given an interactive legal task situated in a dynamic judicial environment, a legal agent plays one of several key judicial roles, such as legal trainee, lawyer, or judge, and engages in multi-round dialogues with other participants in the environment. At each dialogue round, the agent is required to generate an explicit reasoning process and a corresponding response to other roles. The interaction terminates once the task objective is completed, which may involve resolving user confusion, generating a formal legal document, or concluding a judicial procedure.

% During the interaction, the legal agent is allowed to dynamically invoke external legal tools and perform iterative exploration and verification. Formally, at dialogue round $t$, given the dialogue history $\mathcal{H}_t$ and an available tool set $\mathcal{T}$, the agent generates a complete reasoning chain $\mathcal{R}_t$ together with a response $r_t$. We formalize this process as a mapping: $(\mathcal{H}_t, \mathcal{T}) \rightarrow (\mathcal{R}_t, r_t)$.

\begin{figure*}[t]
  \centering
  \includegraphics[width=0.9\linewidth]{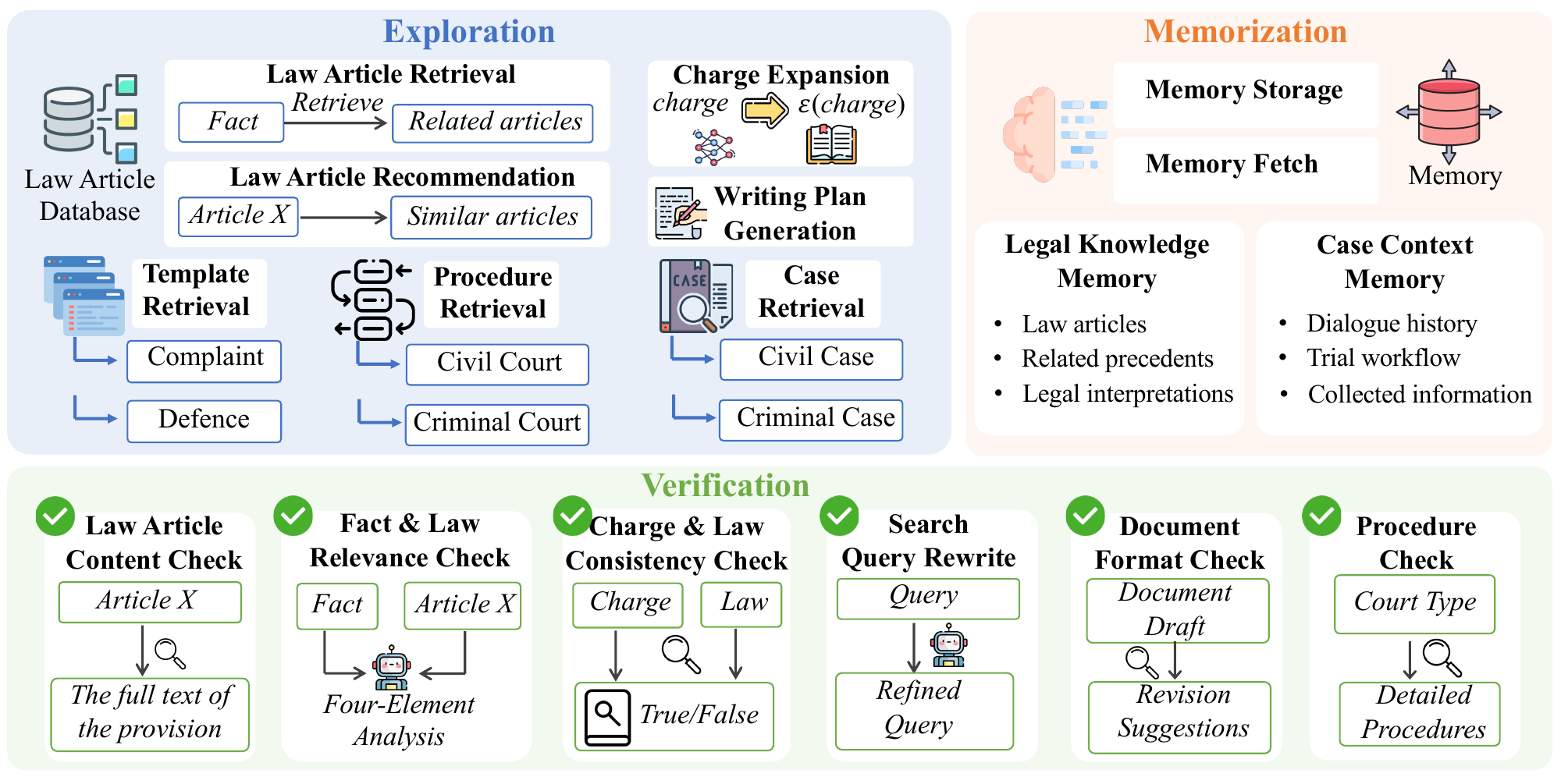}
  \caption{Overview of tools used in exploration, verification, and memorization. Details of these tools can be found in Table~\ref{tab:tool}.}
  \label{fig:tools}
  \Description{}
\end{figure*}

\subsection{Overview of the LawThinker Framework}
\label{subsec: lawthinker framework}
We propose LawThinker, a legal research agent framework that adopts an Explore-Verify-Memorize strategy. The central design principle is to treat knowledge exploration and verification as an enforced atomic operation at the system level. Rather than relying on the model to spontaneously reflect on its own reasoning, LawThinker guarantees that every exploration step is immediately followed by a dedicated verification module. This design prevents errors introduced during knowledge retrieval from propagating into subsequent reasoning steps. An overview of the framework is illustrated in Figure~\ref{fig:pipeline}.

% LawThinker enables agents to autonomously explore external legal knowledge and explicitly verify intermediate reasoning steps, ensuring both factual correctness and procedural compliance throughout the reasoning process. The framework is illustrated in Figure~\ref{fig:pipeline}.

% \begin{figure*}[t]
%   \centering
%   \includegraphics[width=0.9\linewidth]{pic/tool6.pdf}
%   \caption{Overview of tools used in exploration, verification, and memorization. Details of these tools can be found in Table~\ref{tab:tool}.}
%   \label{fig:tools}
%   \Description{}
% \end{figure*}

Given a legal task, the agent begins its reasoning process and invokes exploration tools when it encounters knowledge gaps. The system controller intercepts the result of each exploration and activates a DeepVerifier module, which examines the retrieved information and returns a structured assessment to the main reasoning chain. Based on this feedback, the agent determines whether to proceed with the current information, revise its analysis, or reformulate the query and re-explore. This explore-verify cycle may iterate multiple times within a single dialogue round until sufficient verified knowledge has been gathered, at which point the agent generates the final response. The technical design of the DeepVerifier is detailed in Section~\ref{subsec: deepverifier}.

The framework is built on three critical design decisions. First, the verification step is enforced by the system controller rather than left to the model's discretion. The DeepVerifier is activated after every exploration, ensuring that no retrieved information enters the reasoning chain without examination. Second, verification operates at the granularity of individual reasoning steps. We define a step as the reasoning segment from the initiation of reasoning to the invocation of a single exploration tool. Verifying at this fine granularity allows the system to detect and correct errors before they compound across steps. Third, the framework incorporates a memory module that persistently stores validated legal knowledge and key case context. Both the main reasoning agent and the DeepVerifier can write to memory, ensuring that stored information has passed through verification. The agent retrieves this information in subsequent rounds to avoid redundant exploration. The memory mechanism is described in Section~\ref{subsec: memory mechanism}.

% \begin{itemize}[leftmargin=10pt]
% \item \textbf{Main Reasoning Process:} The core reasoning process interleaves knowledge exploration, deep verification and interaction with a memory module. When encountering knowledge gaps, the LLM autonomously initiates exploration by navigating the legal knowledge base through retrieval or knowledge associations. To ensure reliability, each exploration step is followed by a DeepVerifier module, which performs explicit verification on the retrieved information. This module examines knowledge accuracy, relevance to case facts, and procedural validity, thereby enabling in-depth knowledge acquisition while maintaining a legally sound reasoning trajectory.

% \item \textbf{DeepVerifier Module:} Each knowledge exploration may introduce noise or irrelevant information. To mitigate this risk, LawThinker delegates critical verification to a dedicated DeepVerifier module. The module performs multi-dimensional checks and returns structured checking results, which guide the agent’s correction and reflection in subsequent reasoning steps. To emulate real judicial practice where judges repeatedly consult key legal provisions and precedents, the agent further interacts with a memory module. This interaction supports long-horizon reasoning by enabling persistent storage and reuse of validated information.
% \end{itemize}

\subsection{Legal Knowledge Exploration}
Legal knowledge is not a flat collection of documents but a multi-source, densely connected space. Statutes map to specific charges, related statutes share semantic or structural proximity, and precedents provide empirical references for statutory application. A single retrieval method is insufficient to navigate this interconnected space. To support comprehensive knowledge acquisition, we construct a legal knowledge base consisting of a law article corpus (55,347 provisions covering Civil Law, Criminal Law, and judicial interpretations)~\cite{zhang2025chinese}, a charge corpus (346 standardized criminal charges), a case corpus (criminal and civil judgments from China Judgments Online), and a law-charge mapping dictionary that explicitly links charges to their relevant articles.

Based on these resources, we design seven exploration tools organized into three categories. The design is motivated by the observation that legal knowledge forms a densely connected space rather than a flat document collection. A single retrieval query may locate a relevant statute, but legal reasoning often requires traversing from statutes to related provisions, from provisions to applicable charges, and from charges to precedents. The tools of three categories are designed to support these traversal patterns.

The first category supports statute and charge exploration. It enables the agent to retrieve relevant provisions via dense retrieval (e.g., BGE~\cite{xiao2024c}), discover related articles through semantic and structural similarity, and expand candidate charges by identifying confusing or closely related offenses~\cite{glare}. The second category supports precedent exploration. The agent retrieves similar cases via the SAILER~\cite{li2023sailer} retriever to align its reasoning with established judicial practice. The third category provides task-guided support for structured tasks, including template retrieval and writing plan generation for document drafting, as well as procedure retrieval for courtroom simulation. Detailed descriptions of all tools are provided in Figure~\ref{fig:tools} and Appendix~\ref{Detailed Description of Legal Tools}.

The invocation of these tools is fully autonomous. The agent decides which tools to call and what queries to construct based on the knowledge gaps it perceives during reasoning, without predefined trigger rules or fixed tool sequences.

\subsection{DeepVerifier: Hybrid Step-Level Verification}
\label{subsec: deepverifier}

A natural approach to verifying intermediate reasoning is to let the model reflect on its own output~\cite{madaan2023self,legalreasoner}. However, self-reflection is limited in two ways, and the DeepVerifier is designed to address each of them through specific mechanisms.
First, self-reflection cannot access external ground truth. When the model generates a fabricated statute, it has no means to detect this fabrication because the error originates from its own generation. The DeepVerifier addresses this limitation through grounded verification tools such as Law Article Content Check, which queries authoritative databases to retrieve the full text of a cited provision. This provides a hard factual constraint that is independent of the model's parametric knowledge.
Second, self-reflection operates within the same reasoning context that produced the error. The model tends to treat its previous outputs as established premises and is unlikely to question assumptions already absorbed into the chain.

The DeepVerifier mitigates this by operating under a dedicated verification prompt with a distinct role definition. Although the DeepVerifier shares the same base model, the separated prompt context prevents it from inheriting the reasoning assumptions of the main agent. It evaluates each exploration result from an independent, critical perspective rather than continuing the generative flow. Given the current reasoning step $s_i$, its associated exploration result $e_i$, and the dialogue history $H_t$, the DeepVerifier selects and invokes verification tools along three dimensions that correspond to the core requirements of legal reasoning.

% The DeepVerifier is driven by the same base model as the main reasoning agent but operates under a dedicated verification prompt that defines a distinct role and tool set. As described in Section~\ref{subsec: lawthinker framework}, the system controller activates the DeepVerifier after each exploration step and provides it with three inputs: the current reasoning step $s_i$, the exploration result $e_i$, and the dialogue history $H_t$. The DeepVerifier then selects and invokes appropriate verification tools, and returns a structured verification result to the main reasoning chain. We design six verification tools organized along three dimensions that correspond to the core requirements of legal reasoning.

\textbf{(1) Knowledge Accuracy.} Legal reasoning requires that cited statutes are authentic and complete. LLMs frequently hallucinate statutory content, such as fabricating provisions or mismatching article numbers with their actual text~\cite{yu2025evaluating,guo2025specialized}. To address this, Law Article Content Check directly queries the law article database to retrieve the authoritative full text of a cited provision, providing a hard factual constraint that is independent of the model's reasoning capability. When initial retrieval results are unsatisfactory, Search Query Rewrite refines the original query to improve subsequent exploration quality, ensuring that the agent works with accurate and relevant legal knowledge.

\textbf{(2) Fact-Law Relevance.} A correct statute citation is insufficient if the provision does not apply to the case at hand. Fact-Law Relevance Check analyzes whether cited provisions are applicable to the given case by examining the correspondence between case facts and the four constitutive elements of a crime [31], including subject, object, subjective aspect, and objective aspect. Charge-Law Consistency Check further verifies whether predicted charges are legally supported by the corresponding statutes. Together, these two tools ensure that the mapping between legal knowledge and case facts is logically sound.

\textbf{(3) Procedural Compliance.} Legal tasks often impose strict requirements on the format and sequence of outputs. Procedure Check examines whether the agent follows the correct procedural sequence in courtroom simulation and interacts appropriately with different participants at each stage. Document Format Check analyzes the structural completeness of generated legal documents and provides revision suggestions. These two tools ensure that the agent's outputs conform to legally prescribed norms, covering both judicial procedures and document standards.

The six tools combine two verification strategies. Some tools, such as Law Article Content Check, perform grounded verification by checking information against authoritative databases. Others, such as Fact-Law Relevance Check, perform analytical verification through specialized LLM reasoning with well-defined evaluation criteria. This hybrid design allows the system to eliminate factual errors through hard constraints while handling complex legal judgments that require deeper reasoning. After all checks complete, the structured results are fed back into the main reasoning chain, and the agent autonomously decides whether to proceed, revise its analysis, or initiate a new exploration.

\subsection{Memory Mechanism}
\label{subsec: memory mechanism}
As discussed above,  information in dynamic legal tasks is acquired progressively across dialogue rounds, and the agent must retain verified knowledge for reuse in later rounds. LawThinker addresses this through a memory module that maintains two categories of content.
Legal Knowledge Memory stores validated legal information, including law articles, related charges, precedents, and legal interpretations that have been explored and verified during the reasoning process. Case Context Memory stores task-specific information, including dialogue history, parties' identities, disputed issues, collected evidence, and trial workflow progress. The separation of these two categories reflects their distinct roles: legal knowledge provides the normative basis for reasoning, while case context grounds the reasoning in the specific scenario.

Interaction with the memory module is implemented through two dedicated tools, \verb|memory_store| and \verb|memory_fetch|. The agent autonomously decides what to store and when to retrieve based on its reasoning needs. Both the main reasoning agent and the DeepVerifier can write to memory. The main agent stores task-specific context such as newly identified disputed issues or evidence. The DeepVerifier stores legal knowledge that has been validated during the verification process, such as confirmed statute content or verified fact-law mappings. This dual-channel mechanism ensures that all stored legal knowledge has been examined before entering memory, so that subsequent retrieval does not reintroduce unverified information into the reasoning chain.

\subsection{Legal Agent in Various Scenarios}

% LawThinker can adapt to diverse judicial environments by invoking appropriate tools and performing iterative exploration and verification. Core judicial environments can be categorized into three levels according to task difficulty, interaction complexity, and procedural depth. We next describe the reasoning paradigms of LawThinker across these three representative environments.

The Explore-Verify-Memorize strategy provides a unified reasoning framework, but different judicial scenarios vary significantly in task complexity, interaction depth, and procedural constraints. The framework adapts to these differences through varying tool combinations and verification priorities. We describe three representative levels in order of increasing difficulty.

% LawThinker is designed as a general-purpose legal agent capable of adapting to diverse dynamic judicial scenarios by autonomously selecting appropriate tools and performing iterative exploration and verification. In practice, these scenarios vary substantially in task difficulty, interaction complexity, and procedural depth. Accordingly, we illustrate how LawThinker instantiates its reasoning paradigm across three representative judicial scenarios in J1-ENVS~\cite{j1}.

\textbf{Level I: Knowledge Questioning and Legal Consultation.} At this level, the legal agent acts as a legal trainee and responds to progressive questions from the general public regarding legal knowledge or specific cases. The primary challenge is knowledge accuracy. Users often ask about specific provisions or legal concepts, and any fabricated or misquoted content directly undermines the consultation quality. The agent therefore relies heavily on exploration tools for statute and case retrieval, with verification focused on confirming the authenticity of cited provisions and their relevance to the query.

% At this level, the legal agent acts as a legal trainee and responds to progressive questions posed by the general public regarding legal knowledge or specific cases. Interactions focus on clarifying legal concepts and identifying applicable provisions. LawThinker mainly relies on exploration tools, such as law article retrieval and case retrieval, to access foundational legal knowledge, and applies verification tools to check the correctness of cited provisions and their relevance to the user’s query. This setting emphasizes reliable legal explanation and fact-grounded consultation.

\textbf{Level II: Complaint and Defense Drafting.} At this level, the legal agent acts as a lawyer and engages in multi-turn interactions with litigants to draft formal legal documents. The primary challenge shifts to information completeness and structural compliance. A complaint or defense document must contain specific required components, such as party information, claims, evidence, and legal reasoning, organized in a prescribed format. The agent collects these components across multiple dialogue turns and progressively assembles a complete document. Verification at this level focuses on document format checks to ensure that no required section is missing and the overall structure meets professional standards.

% At this level, the legal agent acts as a lawyer and engages in multi-turn interactions with litigants to draft formal legal documents. Compared with Level I, interactions are more structured, and highly information-intensive. During exploration, LawThinker retrieves writing plans and document templates, follows the plan to gather required information. During verification, the agent performs document format checks to ensure structural completeness and professional compliance and stores key content in memory. This level demonstrates the agent’s capacity to support end-to-end normative drafting in realistic, interactive settings.

\textbf{Level III: Civil and Criminal Courts.} At the highest level, the agent acts as a judge and interacts with multiple parties to conduct a complete courtroom process, ultimately delivering a legally valid judgment. The primary challenge is procedural compliance over a long interaction horizon. A civil trial consists of multiple sequential stages, including preparation, investigation, debate, and mediation, each with mandatory actions that must be executed in the correct order. The agent must track which stages have been completed, ensure that no mandatory step is skipped, and maintain consistency between evolving case facts and the applied legal grounds. Verification at this level relies on procedure checks to monitor stage completion and prevent procedural violations.
%  throughout the proceeding

% At the highest level, the agent acts as a judge and interacts with multiple parties to conduct a complete courtroom process. It follows the full procedural framework of civil or criminal trials and ultimately delivers a legally valid judgment. During exploration, the agent retrieves stage-wise trial procedures and relevant legal knowledge. During verification, it ensures no mandatory procedural step is omitted or executed out of order, and the applied legal grounds remain consistent with the fact, while persistently storing critical contextual information.

% Across these three levels, LawThinker exhibits a unified yet adaptable reasoning paradigm. While the same Explore-Verify-Memorize framework underlies all scenarios, its instantiation varies with interaction patterns and procedural depth. In lightweight consultation tasks, the framework primarily mitigates knowledge hallucination; in drafting and courtroom scenarios, it further ensures structural completeness and procedural legitimacy. This demonstrates LawThinker’s practical suitability for a broad spectrum of dynamic judicial environments.

\section{Experiments}

\subsection{Experimental Setup}

\paragraph{\textbf{Dataset}} 
To evaluate the performance of LawThinker in dynamic and interactive judicial scenarios, we conduct a comprehensive evaluation on J1-EVAL~\cite{j1}, which comprises 508 distinct real-world judicial environments spanning six scenario types, organized into three hierarchical levels according to task complexity and interaction depth. The benchmark covers topics involving diverse entities, multiple types of civil disputes, and a wide range of criminal offense categories, enabling an effective assessment of LawThinker’s adaptability across different legal contexts.

% \textbf{\textit{Level-I}} focuses on Knowledge Questioning and Legal Consultation; \textbf{\textit{Level-II}} covers Complaint Drafting and Defence Drafting; and \textbf{\textit{Level-III}} involves Civil Court and Criminal Court. This hierarchical design enables a fine-grained evaluation of legal reasoning performance as task complexity and interaction depth increase.

\paragraph{\textbf{Metrics}}
For evaluation metrics, we adopt the same measures as those used in J1-EVAL. Given the varying complexity across judicial environments, it adopts fine-grained, scenario-specific evaluations, combining \textit{rule-based} and \textit{LLM-based} approaches. For Level-I, which includes both binary questions and open-ended queries, we report the average score across these two question types. For Level-II, the Format-Following Score (FOR) assesses adherence to document structure, focusing on the ordering and presence of required components, while the Document Score (DOC) evaluates the accuracy and completeness of the content, such as the information of plaintiff and defendant, claims, and supporting evidence. For Level-III, the Procedural-Following Score (PFS) measures the completeness of different stages in court proceedings; the Civil Judgment Score (JUD) evaluates the accuracy of judicial decisions; Crime Accuracy (CRI) assesses the correctness of predicted charges; VER measures the accuracy of fines and sentences using log-distance; and Law Accuracy (LAW) evaluates the completeness and correctness of cited legal provisions. Overall, these metrics can be broadly categorized into \textbf{outcome-oriented} and \textbf{process-oriented} evaluations, enabling a comprehensive assessment of both final results and intermediate reasoning processes.

\paragraph{\textbf{Baselines}}
We compare our method against three categories of baseline approaches. \textbf{\textit{(1) Direct Reasoning:}} We employ different types of models to drive the legal agent, including both general-purpose multilingual LLMs and legal-specific LLMs: Qwen2.5-7B-Instruct~\cite{yang2025qwen2}, Qwen3-8B/14B/32B~\cite{yang2025qwen3}, Ministral-8B-Instruct-2410~\cite{chaplot2023albert}, GLM-4-9B-Chat~\cite{glm2024chatglm}, Gemma3-12B-IT~\cite{gemma_2025}; ChatLaw2-7B~\cite{cui2023chatlaw}, and LawLLM-7B~\cite{yue2023disclawllm,yue2024lawllm}. \textbf{\textit{(2) Workflow-based Methods:}} ReAct~\cite{yao2022react} interleaves reasoning with task-specific actions, Plan-and-Solve~\cite{wang2023plan} first decomposes the overall task into a sequence of manageable subtasks and then solves them according to a predefined plan, and Plan-and-Execute~\cite{topsakal2023creating} formulates a multi-step plan that is executed sequentially, with the model revisiting and refining the plan upon task completion. \textbf{\textit{(3) Autonomous Tool Usage within Reasoning:}} Search-o1~\cite{li2025search} augments reasoning by dynamically retrieving external knowledge when the model encounters uncertain or ambiguous information. For a fair comparison, we provide all baseline methods with access to exploration tools.

\begin{table*}[t]
  \centering
  \small
  \caption{Performance comparisons on J1-EVAL. KQ: Knowledge Questioning, LC: Legal Consultation, CD: Complaint Drafting, DD: Defence Drafting, CI: Civil Court, CR: Criminal Court. Best results are in bold and ``-'' denotes failure to complete the task.}
  \begin{tabular}{p{0.17\textwidth}ccccccccccccccc}
    \toprule
    \multirow{3}{*}{\textbf{Methods}} 
    & \multicolumn{2}{c}{\textbf{Level-I}} & \multicolumn{4}{c}{\textbf{Level-II}}
    & \multicolumn{9}{c}{\textbf{Level-III}} \\
    \cmidrule(lr){2-3} \cmidrule(lr){4-7} \cmidrule(lr){8-16}
    & \multicolumn{1}{c}{\textbf{KQ}} & \multicolumn{1}{c}{\textbf{LC}} 
    & \multicolumn{2}{c}{\textbf{CD}} & \multicolumn{2}{c}{\textbf{DD}}
    & \multicolumn{4}{c}{\textbf{CI}} & \multicolumn{5}{c}{\textbf{CR}} \\
    \cmidrule(lr){2-2} \cmidrule(lr){3-3} \cmidrule(lr){4-5} \cmidrule(lr){6-7} \cmidrule(lr){8-11} \cmidrule(lr){12-16}
    & \textbf{Avg.} & \textbf{Avg.} & \textbf{FOR} & \textbf{DOC} & \textbf{FOR} & \textbf{DOC} & \textbf{PFS} & \textbf{JUD} & \textbf{REA} & \textbf{LAW} & \textbf{PFS} & \textbf{CRI} & \textbf{VER} & \textbf{REA} & \textbf{LAW} \\
    \midrule
    \multicolumn{15}{l}{\textit{\textbf{Direct Reasoning}}} \\
    \multicolumn{15}{l}{\textit{Multilingual LLMs}} \\
    Qwen2.5-7B & 52.9 & 40.1 & 73.1 & 76.9 & 26.5 & 50.4 & 23.4 & 9.1 & 20.0 & 8.5 & 22.4 & 81.2 & 68.1 & 54.8 & 23.9 \\
    Ministral-8B & 49.1 & 32.1 & 50.5 & 65.3 & 40.1 & 52.8 & 15.1 & 5.7 & 10.1 & 4.3 & 5.7 & 11.6 & 16.2 & 7.4 & 2.7 \\
    Qwen3-8B & 52.8 & 40.3 & 7.5 & 57.9 & 12.2 & 40.7 & 12.9 & 9.9 & 23.1 & 14.7 & 22.0 & 49.3 & 46.5 & 35.1 & 25.9 \\
    GLM4-9B & 55.6 & 34.2 & 52.5 & 67.0 & 57.0 & 54.8 & 20.8 & 5.9 & 16.5 & 4.7 & 31.4 & 66.7 & 54.4 & 40.3 & 17.5 \\
    Gemma3-12B & 51.2 & 33.5 & 57.2 & 77.0 & 11.8 & 51.6 & 42.7 & 17.7 & 32.8 & 14.8 & 33.2 & 72.5 & 61.3 & 44.5 & 18.5 \\
    Qwen3-14B & 57.7 & 43.3 & 69.3 & \textbf{77.5} & 32.6 & 59.0 & 6.8 & 4.1 & 5.9 & 3.9 & 30.7 & 55.1 & 49.6 & 44.8 & 30.8 \\
    Qwen3-32B & 59.6 & 47.4 & 71.4 & 66.4 & 30.8 & 51.0 & 42.5 & 16.7 & 35.7 & 17.3 & 33.7 & 69.6 & 57.8 & 55.5 & 33.9 \\
    % DeepSeek-R1-671B &  &  &  &  &  &  &  &  &  &  &  &  &  &  &  \\
    \multicolumn{15}{l}{\textit{Legal-specific LLMs}} \\
    ChatLaw2-7B & 52.9 & 36.5 & 30.3 & 24.0 & 15.8 & 18.9 & 3.7 & 5.6 & 7.8 & - & 3.7 & 24.6 & 20.9 & 11.9 & 2.2 \\
    LawLLM-7B & 48.7 & 35.3 & 74.2 & 69.9 & 26.5 & 43.6 & 3.4 & - & - & - & 2.2 & 13.0 & 7.9 & 10.0 & 1.6 \\
    \midrule
    \multicolumn{15}{l}{\textit{\textbf{Workflow-based Methods (with Qwen3-32B)}}} \\
    ReAct & 61.3 & 46.8 & 50.1 & 67.5 & 41.2 & 54.5 & 39.2 & 21.6 & 38.3 & 20.8 & 24.3 & 75.4 & 62.1 & 58.8 & 34.5 \\
    Plan-and-Solve & 60.9 & 46.4 & 53.3 & 68.5 & 42.7 & 59.0 & 38.0 & 23.4 & 38.7 & 19.4 & 23.8 & 79.7 & 65.3 & 57.7 & 36.6 \\
    Plan-and-Execute & 61.9 & 47.9 & 53.1 & 62.8 & 40.1 & 55.4 & 40.5 & 23.8 & 42.6 & 20.5 & 24.2 & 75.4 & 57.9 & 58.3 & 34.5 \\
    \midrule
    \multicolumn{15}{l}{\textit{\textbf{Autonomous Tool Usage within Reasoning (with Qwen3-32B)}}} \\
    Search-o1 & 63.5 & 48.4 & 53.3 & 70.3 & 42.7 & 54.9 & 40.6 & 20.5 & 36.3 & 21.2 & 28.9 & 84.1 & 67.1 & 60.9 & 39.5 \\
    \rowcolor[RGB]{236,244,252}
    LawThinker (ours) & \textbf{64.1} & \textbf{52.4} & \textbf{87.7} & 75.6 & \textbf{81.4} & \textbf{60.3} & \textbf{50.3} & \textbf{28.5} & \textbf{44.0} & \textbf{23.2} & \textbf{41.5} & \textbf{84.1} & \textbf{69.7} & \textbf{63.9} & \textbf{40.4} \\
    \bottomrule
  \end{tabular}
  \label{tab:main result}
\end{table*}

\paragraph{\textbf{Implementation Details}} 
We use Qwen3-32B to drive the environment, where it plays the role of non-critical agents such as general public, plaintiffs and defendants. To avoid unbounded interaction loops due to model limitations while ensuring sufficient conversational coverage, we impose an upper bound on the number of interaction rounds for each scenario: 15 for knowledge questioning, 10 for legal consultation, 30 for complaint drafting, 30 for defence drafting, 60 for civil court, and 50 for criminal court. For LLM-based metric evaluation, we adopt GPT-4o~\cite{achiam2023gpt}. All experiments are conducted on two NVIDIA A800-80GB GPUs.

\subsection{Main Result}
\paragraph{\textbf{Overall Performance}} The main results are presented in Table~\ref{tab:main result}, demonstrating the superiority of LawThinker in dynamic judicial environments. More detailed metrics are reported in Appendix~\ref{Additional overall performance}.

\textbf{(1) Direct Reasoning exhibits consistently limited performance across all tasks}, with particularly poor results in more complex scenarios. This is largely due to severe hallucination issues in legal reasoning, such as fabricating legal provisions. Since direct reasoning relies solely on the model’s internal knowledge, its overall accuracy remains below 50\%. In Level-I, users progressively query specific legal knowledge or case details, placing high demands on comprehensive legal understanding. Models fail to explore sufficiently broad legal knowledge and lack mechanisms to verify intermediate reasoning steps, leading to unreliable conclusions. Level-II and Level-III further require procedural awareness, yet models struggle to follow legally grounded procedures, yielding low scores on process-oriented metrics. This limitation is particularly evident in legal-specific LLMs such as ChatLaw and LawLLM, as weaker interaction abilities and limited long-horizon contextual reasoning hinder their performance in dynamic settings.

\textbf{(2) Workflow-based Methods outperform direct reasoning by incorporating external knowledge}, but their gains are constrained by the absence of explicit verification mechanisms, which often introduce noise. Approaches such as ReAct interleave reasoning with actions to acquire richer legal information, partially mitigating hallucinations and achieving over a 13\% improvement in overall accuracy compared to direct reasoning. However, without systematic checking, these methods exhibit degraded procedural compliance. Notably, in the Complaint Drafting scenario, their performance can even underperform direct reasoning, highlighting the necessity of verifying intermediate reasoning steps rather than simply increasing access to external information.

\begin{table}[t]
  \centering
  \small
  \caption{Accuracy evaluation on different legal benchmarks.}    \label{static}
  % \begin{tabular}{lcccc}
  \setlength\tabcolsep{3pt}
  \begin{tabular}{lcccc}%>{\raggedright\arraybackslash}p{2.1cm}>{\centering\arraybackslash}p{1.2cm}>{\centering\arraybackslash}p{1.2cm}>{\centering\arraybackslash}p{1.3cm}>{\centering\arraybackslash}p{1cm}}
    \toprule
    \textbf{Method} & \textbf{LawBench} & \textbf{LexEval} & \textbf{Unilaw-R1-Eval} & \textbf{Avg.(\%)} \\
    \midrule
    Direct Reasoning & 52.0 & 67.1 & 48.3 & 55.8 \\
    ReAct & 56.2 & 68.6 & 51.3 & 58.7 \\
    Plan-and-Solve & 57.9 & 66.6 & 50.3 & 58.3 \\
    Plan-and-Execute & 57.0 & 68.1 & 52.0 & 59.0 \\
    Search-o1 & 55.3 & 69.0 & 49.7 & 58.0 \\
    \rowcolor[RGB]{236,244,252}
    LawThinker & \textbf{62.0} & \textbf{71.1} & \textbf{53.7} & \textbf{62.3} \\
    \bottomrule
  \end{tabular}
  
\end{table}

\begin{figure}[t]
  \centering
  \includegraphics[width=0.9\linewidth]{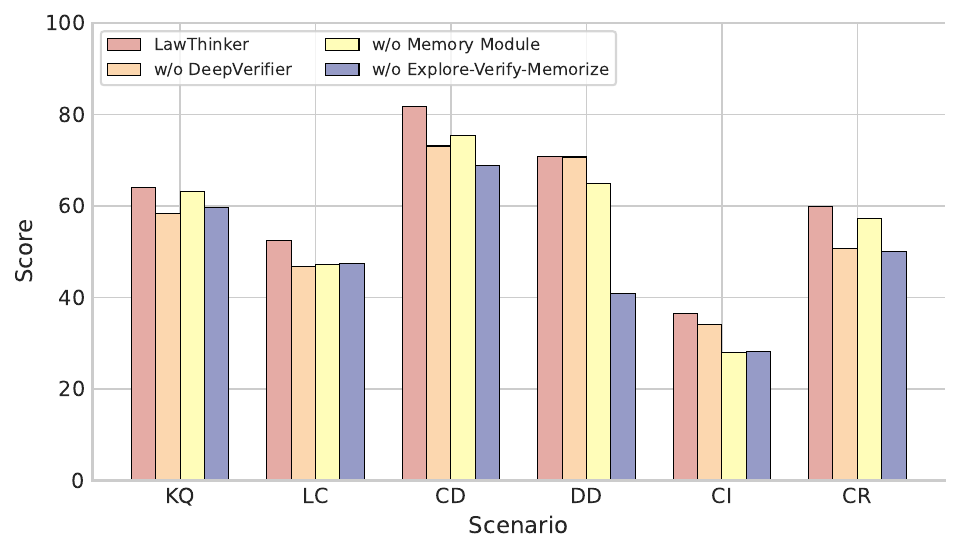}
  \caption{Ablation study across six legal scenarios.}
  \label{fig:ablation}
  \Description{}
\end{figure}

\textbf{(3) LawThinker consistently outperforms all baselines across scenarios}, validating the effectiveness of the Explore-Verify-Memorize strategy. Overall, LawThinker achieves a 24\% improvement over direct reasoning and an 11\% gain over workflow-based methods, with particularly strong performance on process-oriented metrics such as FOR and PFS, highlighting the value of step-level verification for improving process accuracy and procedural compliance. Search-o1 performs an in-depth analysis of retrieved documents, but its reasoning is mainly based on self-reflection, with limited support from verified external knowledge. We observe that Defence Drafting is more challenging than Complaint Drafting, as it requires comprehensively gathering relevant information and constructing more complex counterarguments in response to the opposing party’s claims. In Level-III, LawThinker more effectively completes each stage of court proceedings and attains higher accuracy in judgment generation, legal reasoning, and statute citation. Overall, LawThinker demonstrates robust performance across judicial environments ranging from inquiry to adjudication, offering promising evidence toward the development of reliable and comprehensive legal intelligence.

% \begin{figure}[t]
%   \centering
%   \includegraphics[width=0.9\linewidth]{pic/ablation2.pdf}
%   \caption{Ablation study across six legal scenarios.}
%   \label{fig:ablation}
%   \Description{}
% \end{figure}

\paragraph{\textbf{Generalization to Static Scenarios}} To assess the generalization ability of LawThinker in static legal scenarios, we evaluate it on three Chinese legal domain benchmarks: LexEval~\cite{li2024lexeval}, LawBench~\cite{fei2024lawbench}, and UniLaw-R1-Eval~\cite{cai2025unilaw}. Detailed descriptions of these benchmarks are provided in Appendix~\ref{Descriptions of Static Benchmarks}, and evaluation is conducted using the original metrics defined by each benchmark. We conduct comparative experiments between LawThinker and other reasoning methods based on the Qwen3-32B model. As shown in Table~\ref{static}, LawThinker consistently achieves the best performance across all benchmarks, delivering an average accuracy improvement of approximately 6\% over direct reasoning baselines. These results further validate the effectiveness of the proposed Explore-Verify-Memorize strategy for legal tasks. By autonomously invoking specialized legal tools, the model is able to acquire comprehensive legal knowledge, while the DeepVerifier module ensures the accuracy of each reasoning step and adherence to legal procedures. Overall, our approach aligns well with the requirements of legal practice, where both static and dynamic scenarios demand a high degree of rigor and reliability in the reasoning process.

\subsection{Ablation Study}
% \begin{figure}[t]
%   \centering
%   \includegraphics[width=0.9\linewidth]{pic/ablation2.pdf}
%   \caption{Ablation study across six legal scenarios.}
%   \label{fig:ablation}
%   \Description{}
% \end{figure}

\begin{figure}[t]
  \centering
  \includegraphics[width=0.9\linewidth]{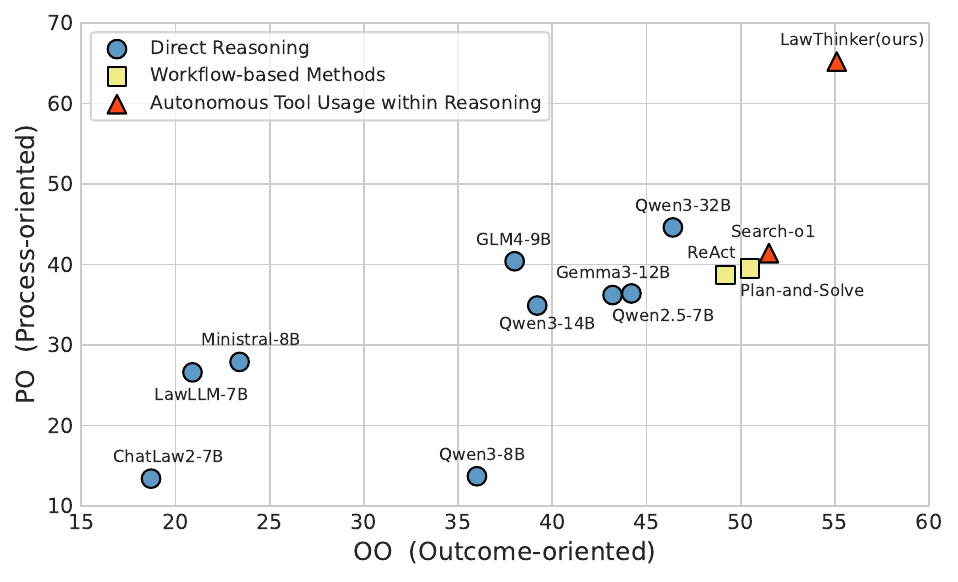}
  \caption{Performance with outcome-oriented and process-oriented metrics across models and reasoning paradigms.}
  \label{fig:oo vs po}
  \Description{}
\end{figure}

To examine the contributions of exploration, verification, and memorization in LawThinker, we conduct ablation studies across six scenarios. The results are shown in Figure~\ref{fig:ablation}.

\textbf{\textit{(1) DeepVerifier is critical to overall performance.}} Removing DeepVerifier leads to consistent performance drops across all scenarios, indicating that verifying intermediate reasoning steps is essential for legal tasks. This is particularly pronounced in knowledge-intensive scenarios such as KQ and LC, as well as in CR, where strict procedural compliance is required. These results show that explicit verification effectively reduces hallucinations and reasoning errors.

\textbf{\textit{(2) The memory module substantially benefits long-horizon tasks.}} In document drafting and court simulation scenarios, the agent must retain information gathered across multiple turns to generate final documents or judgments. Consequently, removing the memory module causes more severe degradation in these tasks. In contrast, KQ relies mainly on information from the current turn and is therefore less sensitive to historical memory.

\textbf{\textit{(3) The Explore-Verify-Memorize strategy yields the largest gains.}} When exploration, verification, or memorization is absent, the model fails to acquire accurate external knowledge and cannot support long-horizon reasoning, leading to performance degradation across all scenarios compared to LawThinker. This confirms that the synergy of exploration, verification, and memorization is crucial for robust legal reasoning.

\subsection{Quantitative Analysis}
To comprehensively evaluate the performance of LawThinker, we conduct multiple quantitative experiments, analyzing both outcome accuracy and procedural compliance.

% \begin{figure}[t]
%   \centering
%   \includegraphics[width=0.9\linewidth]{pic/OO_PO_comparison2.pdf}
%   \caption{Performance with outcome-oriented and process-oriented metrics across models and reasoning paradigms.}
%   \label{fig:oo vs po}
%   \Description{}
% \end{figure}

\begin{figure}[!t]
  \centering
  \includegraphics[width=\linewidth]{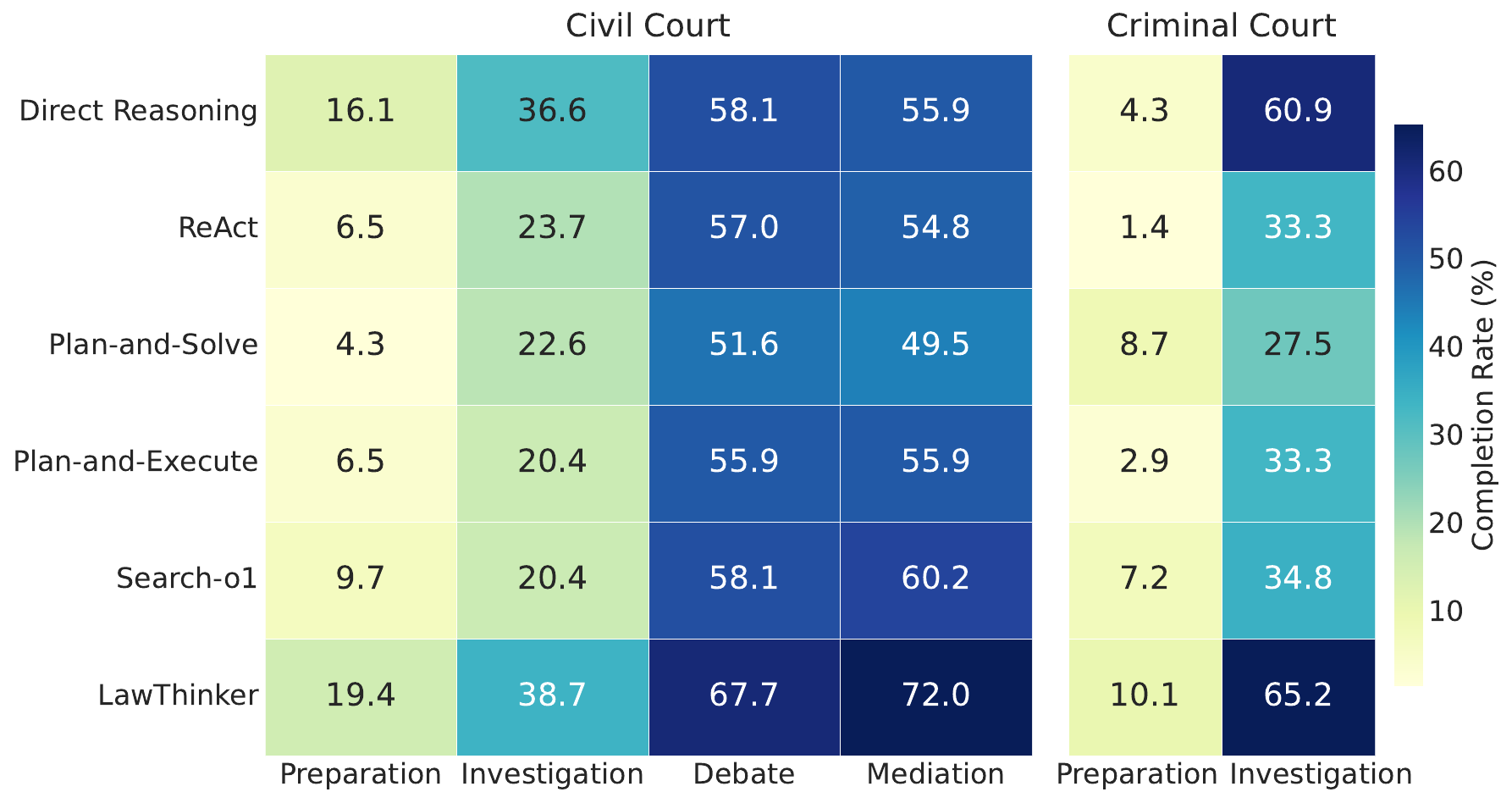}
  \caption{Court-stage completion rates across different methods. Left: Civil Court. Right: Criminal Court.}
  \label{fig:heatmap}
  \Description{}
\end{figure}

\paragraph{\textbf{Comprehensive analysis of outcome and process performance.}} In legal reasoning tasks, the correctness of final outcomes is as critical as the compliance of the reasoning process. We evaluate the performance of different models and methods along these two dimensions, as illustrated in Figure~\ref{fig:oo vs po}. LawThinker consistently demonstrates strong performance in both OO and PO, indicating that it effectively improves not only decision accuracy but also procedural compliance throughout the reasoning chain. Within the same reasoning paradigm, larger models generally achieve higher OO and PO scores. However, even large-scale models exhibit constrained PO performance in the absence of explicit verification. Although workflow-based methods often attain higher OO scores by incorporating external knowledge, their PO scores can fall below those of direct reasoning, suggesting that introducing unverified external information may enhance outcome accuracy while undermining procedural correctness in complex legal settings.

\paragraph{\textbf{Analysis of court process completeness.}} To examine procedural completeness across methods in civil and criminal court settings, we compute stage completion rates and visualize them using heatmaps, as shown in Figure~\ref{fig:heatmap}. LawThinker achieves the highest scores across all four civil stages and both criminal stages, reflecting its strong ability to adhere to judicial procedures. In contrast, most other methods score below 25\% in the Preparation and Investigation stages, highlighting a lack of explicit control in fine-grained processes such as information gathering and evidence verification. Notably, all methods exceed 50\% completion in the Debate stage, indicating that argumentative exchanges around disputed issues are relatively easier to generate. Moreover, most approaches perform better in criminal Investigation than in civil cases, likely due to the higher degree of standardization in criminal procedures, which reduces the likelihood of omissions. Interestingly, workflow-based methods such as ReAct underperform direct reasoning, underscoring that exploration without verification is insufficient to ensure procedural compliance. Overall, these results emphasize the critical role of deep verification in improving judicial process completeness, and LawThinker demonstrates clear and consistent advantages across both civil and criminal court scenarios.

% \begin{figure}[!t]
%   \centering
%   \includegraphics[width=\linewidth]{pic/heatmap3.pdf}
%   \caption{Completion rates of court stages.}
%   \label{fig:heatmap}
%   \Description{}
% \end{figure}

\section{Conclusion}
In this work, we propose LawThinker, an autonomous legal research agent designed for dynamic judicial environments. LawThinker adopts an Explore-Verify-Memorize strategy that integrates iterative knowledge exploration with explicit verification. We further design 15 specialized tools spanning exploration, verification, and memorization. Extensive experiments on a dynamic legal benchmark and three static benchmarks demonstrate that LawThinker significantly outperforms existing methods, achieving both accurate outcomes and legally compliant reasoning processes.

% In this work, we propose LawThinker, an autonomous legal research agent that integrates iterative exploration with explicit verification throughout the reasoning process. To equip LawThinker with deep verification capabilities, we introduce a DeepVerifier module to examine each exploration step and ensure the accuracy and procedural compliance of the reasoning process. We also design 15 general-purpose tools tailored for real-world legal scenarios, covering three core dimensions: exploration, verification, and memorization. Extensive experiments on a dynamic legal benchmark and three static benchmarks demonstrate the effectiveness of the Explore-Verify-Memorize strategy. Overall, our framework realizes comprehensive legal intelligence, spanning legal question answering, document drafting, and judgment generation.

% \bibliograpperrefstyle{ACM-Reference-Format}
\bibliographystyle{ACM-Reference-Format}
\bibliography{main}

\clearpage
\newtcolorbox{promptbox}[1][]{
  colback=gray!5!white,
  colframe=gray!75!black,
  fonttitle=\bfseries,
  left=5pt,
  right=5pt,
  top=5pt,
  bottom=5pt,
  breakable=false,
  enhanced,
  #1
}
\appendix

\begin{table*}[t]
\centering
\small
\setlength{\tabcolsep}{6pt}
\caption{Overview of exploration, verification, and memorization tools used by the legal agent.}
\label{tab:tool}

\begin{tabular}{
p{3.5cm}
p{3.7cm}
p{3.5cm}
p{2.0cm}
p{2.3cm}
}
\toprule
\textbf{Tool Name} &
\textbf{Description} &
\textbf{Function} &
\textbf{Inputs} &
\textbf{Outputs} \\
\midrule

\rowcolor{gray!20}
\multicolumn{5}{l}{\textbf{Exploration Tools}} \\

Law Article Retrieval &
Retrieve relevant law articles. &
\texttt{law\_retrieval} &
Query; top-$k$ &
Law articles \\

Law Article Recommendation &
Recommend similar law articles. &
\texttt{law\_recommendation} &
Law article &
Similar laws \\

Charge Expansion &
Expand related criminal charges. &
\texttt{charge\_expansion} &
Charges; top-$k$ &
Expanded charges \\

Case Retrieval &
Retrieve similar cases. &
\texttt{case\_retrieval} &
Case type; query &
Similar cases \\

Template Retrieval &
Retrieve document templates. &
\texttt{template\_retrieval} &
Template type &
Document template \\

Writing Plan Generation &
Generate document writing plan. &
\texttt{plan\_generation} &
Document type &
Writing plan \\

Procedure Retrieval &
Retrieve court procedures. &
\texttt{procedure\_retrieval} &
Court type; stage &
Procedural steps \\

\midrule
\rowcolor{gray!20}
\multicolumn{5}{l}{\textbf{Checking Tools}} \\

Law Article Content Check &
Verify law article content. &
\texttt{law\_check} &
Law name; &
Verified content \\

Fact \& Law Relevance Check &
Check law applicability to facts. &
\texttt{fact\_law\_relevance\_check} &
Case facts; law &
Relevance result \\

Charge \& Law Consistency Check &
Check charge-law consistency. &
\texttt{crime\_law\_consistency\_check} &
Charge; law &
Consistency result \\

Search Query Rewrite &
Rewrite law retrieval queries. &
\texttt{law\_query\_rewrite} &
Query; context &
Refined query \\

Document Format Check &
Check document format. &
\texttt{document\_format\_check} &
Document type; document &
Format feedback \\

Procedure Check &
Check procedural compliance. &
\texttt{procedure\_check} &
Court type &
Procedural guidance \\

\midrule
\rowcolor{gray!20}
\multicolumn{5}{l}{\textbf{Memory Tools}} \\

Memory Storage &
Store knowledge or context. &
\texttt{storeMemory} &
Memory type; content &
Storage status \\

Memory Fetch &
Retrieve stored memory. &
\texttt{fetchMemory} &
Memory type &
Retrieved memory \\

\bottomrule
\end{tabular}

\end{table*}

\begin{table*}[t]
  \centering
  \caption{Fine-grained evaluation at Level I and Level II in J1-EVAL.}
  \begin{tabular}{lcccccccccccc}
    \toprule
    \multirow{3}{*}{\textbf{Methods}} 
    & \multicolumn{4}{c}{\textbf{Level-I}} & \multicolumn{8}{c}{\textbf{Level-II}} \\
    \cmidrule(lr){2-5}\cmidrule(lr){6-13}
    & \multicolumn{2}{c}{\textbf{KQ}} & \multicolumn{2}{c}{\textbf{LC}} & \multicolumn{5}{c}{\textbf{CD}} & \multicolumn{3}{c}{\textbf{DD}} \\
    \cmidrule(lr){2-3} \cmidrule(lr){4-5} \cmidrule(lr){6-10} \cmidrule(lr){11-13}
    & BIN & NBIN & BIN & NBIN & PLA & DEF & CLA & EVI & F\&R & RES & DEF & EVI \\
    \midrule
    \multicolumn{13}{l}{\textit{\textbf{Direct Reasoning}}} \\
    \multicolumn{13}{l}{\textit{Multilingual LLMs}} \\
    Qwen2.5-7B & 56.6 & 49.3 & 44.1 & 36.0 & 91.2 & 91.8 & 76.9 & 48.1 & 76.8 & 67.2 & 62.2 & 21.7 \\
    Ministral-8B & 56.2 & 41.9 & 48.5 & 15.7 & 82.7 & 75.4 & 77.8 & 36.3 & 54.0 & 75.4 & 54.9 & 28.0 \\
    Qwen3-8B & 53.3 & 52.4 & 45.8 & 34.7 & 68.3 & 68.1 & 63.9 & 29.0 & 60.1 & 57.3 & 50.5 & 14.1 \\
    GLM4-9B & 64.0 & 47.1 & 43.4 & 25.0 & 80.9 & 72.5 & 80.0 & 37.2 & 64.4 & 72.9 & 64.0 & 27.5 \\
    Gemma3-12B & 60.0 & 42.4 & 48.0 & 19.0 & 92.2 & 84.2 & 89.6 & 50.9 & 68.1 & 68.9 & 72.4 & 13.4 \\
    Qwen3-14B & 58.3 & 57.0 & 50.9 & 35.7 & 91.2 & 90.6 & 89.2 & 54.6 & 62.0 & 76.3 & 75.8 & 24.9 \\
    Qwen3-32B & 60.7 & 58.5 & 51.1 & 43.7 & 70.6 & 79.6 & 77.6 & 52.5 & 51.6 & 66.1 & 69.6 & 17.4 \\
    \multicolumn{13}{l}{\textit{Legal-specific LLMs}} \\
    ChatLaw2-7B & 57.8 & 47.9 & 51.5 & 21.5 & 32.6 & 28.6 & 27.5 & 11.1 & 20.2 & 20.6 & 26.9 & 9.2 \\
    LawLLM-7B & 55.0 & 42.3 & 51.5 & 19.1 & 84.9 & 82.6 & 75.4 & 44.0 & 62.6 & 59.6 & 54.3 & 17.0 \\
    \midrule
    \multicolumn{13}{l}{\textit{\textbf{Workflow-based Methods}}} \\
    ReAct-Qwen3-32B & 57.6 & 65.0 & 55.0 & 38.6 & 76.6 & 79.1 & 75.2 & 50.1 & 56.6 & 72.6 & 63.5 & 27.4 \\
    Plan-and-Solve-Qwen3-32B & 59.0 & 62.8 & 45.9 & 46.8 & 80.1 & 78.9 & 77.2 & 49.5 & 57.1 & 77.2 & 70.9 & 29.0 \\
    Plan-and-Execute-Qwen3-32B & 60.2 & 63.7 & 53.0 & 42.8 & 71.3 & 72.3 & 72.3 & 44.2 & 54.1 & 73.6 & 66.7 & 26.0 \\
    \midrule
    \multicolumn{13}{l}{\textit{\textbf{Autonomous Tool Usage within Reasoning}}} \\
    Search-o1-Qwen3-32B & 63.8 & 63.3 & 54.4 & 43.2 & 81.9 & 84.9 & 81.0 & 45.3 & 58.2 & 72.6 & 63.5 & 27.4 \\
    LawThinker-Qwen3-32B(ours) & 63.4 & 64.7 & 58.0 & 46.8 & 84.5 & 86.3 & 88.0 & 50.0 & 69.0 & 82.7 & 71.5 & 26.7 \\
    \bottomrule
  \end{tabular}
  \label{tab:result1}
\end{table*}

\begin{table*}[t]
  \centering
  \caption{Fine-grained evaluation at Level III in J1-EVAL.}
  % \begin{tabular}{lcccccccc}
  \begin{tabular}{>{\raggedright\arraybackslash}p{5.5cm}>{\centering\arraybackslash}p{0.8cm}>{\centering\arraybackslash}p{0.8cm}>{\centering\arraybackslash}p{0.8cm}>{\centering\arraybackslash}p{0.8cm}>{\centering\arraybackslash}p{0.8cm}>{\centering\arraybackslash}p{0.8cm}>{\centering\arraybackslash}p{0.8cm}>{\centering\arraybackslash}p{0.8cm}}
    \toprule
    \multirow{3}{*}{\textbf{Methods}} 
    & \multicolumn{8}{c}{\textbf{Level-III}} \\
    \cmidrule(lr){2-9}
    & \multicolumn{3}{c}{\textbf{CI}} & \multicolumn{5}{c}{\textbf{CR}} \\
    \cmidrule(lr){2-4} \cmidrule(lr){5-9}
    & STA & ACT & UNI & STA & ACT & UNI & SEN & FINE \\
    \midrule
    \multicolumn{9}{l}{\textit{\textbf{Direct Reasoning}}} \\
    \multicolumn{9}{l}{\textit{Multilingual LLMs}} \\
    Qwen2.5-7B & 13.4 & 33.3 & 48.4 & 8.0 & 36.9 & 87.0 & 73.6 & 62.7 \\
    Ministral-8B & 13.2 & 17.0 & 21.5 & 2.9 & 8.5 & 14.5 & 20.0 & 12.4 \\
    Qwen3-8B & 9.4 & 16.3 & 43.0 & 21.0 & 23.0 & 59.4 & 54.2 & 38.9 \\
    GLM4-9B & 19.4 & 22.2 & 41.9 & 31.2 & 31.7 & 75.4 & 61.2 & 47.7 \\
    Gemma3-12B & 36.0 & 49.4 & 64.5 & 29.0 & 37.5 & 91.3 & 70.4 & 52.1 \\
    Qwen3-14B & 5.1 & 8.4 & 11.8 & 21.0 & 40.4 & 60.9 & 51.3 & 47.9 \\
    Qwen3-32B & 37.6 & 47.4 & 62.4 & 30.4 & 37.1 & 76.8 & 58.2 & 57.5 \\
    \multicolumn{9}{l}{\textit{Legal-specific LLMs}} \\
    ChatLaw2-7B & 2.2 & 5.2 & 29.0 & - & 7.4 & 33.3 & 25.5 & 16.3 \\
    LawLLM-7B & 3.2 & 3.6 & 4.3 & 1.4 & 2.9 & 15.9 & 12.4 & 3.5 \\
    \midrule
    \multicolumn{9}{l}{\textit{\textbf{Workflow-based Methods}}} \\
    ReAct-Qwen3-32B & 33.3 & 45.1 & 67.7 & 16.7 & 31.9 & 79.7 & 62.5 & 61.7 \\
    Plan-and-Solve-Qwen3-32B & 30.9 & 45.0 & 72.0 & 13.8 & 33.7 & 88.4 & 66.5 & 64.2 \\
    Plan-and-Execute-Qwen3-32B & 33.1 & 47.9 & 73.1 & 16.7 & 31.7 & 78.3 & 58.9 & 56.8 \\
    \midrule
    \multicolumn{9}{l}{\textit{\textbf{Autonomous Tool Usage within Reasoning}}} \\
    Search-o1-Qwen3-32B & 34.9 & 46.3 & 71.0 & 17.4 & 40.4 & 91.3 & 70.0 & 64.1 \\
    LawThinker-Qwen3-32B(ours) & 44.6 & 55.9 & 72.0 & 33.3 & 49.7 & 91.3 & 70.1 & 69.4 \\
    \bottomrule
  \end{tabular}
  \label{tab:result2}
\end{table*}

\section{Detailed Description of Legal Tools} \label{Detailed Description of Legal Tools}
To support long-horizon legal reasoning and sustained interaction across diverse judicial scenarios, the legal agent is equipped with a suite of specialized tools spanning exploration, verification, and memorization. These tools enable the agent to actively retrieve and expand relevant legal knowledge, examine the validity and consistency of intermediate reasoning steps, and preserve contextual information across multi-turn interactions. Table~\ref{tab:tool} provides an overview of the tool taxonomy, including their core functions, inputs, and outputs.

\section{Additional overall performance} \label{Additional overall performance}
We provide additional results of the overall performance in J1-EVAL. In Level I, Binary Accuracy (BIN) measures correctness on binary questions that require a yes-or-no answer, while Non-Binary Accuracy (NBIN) evaluates the accuracy and completeness of responses to open-ended questions. In Level II, the quality of generated legal documents is assessed based on key components. For complaint documents, the evaluation covers information about the plaintiff (PLA) and the defendant (DEF), claims (CLA), evidence (EVI), as well as facts and reasoning (F\&R). Similarly, defence documents are evaluated based on respondent information (RES), defence arguments (DEF), and supporting evidence (EVI). In Level III, completion of a court stage requires the execution of all mandatory procedural actions. Specifically, civil trials are organized into five sequential stages: preparation, investigation, debate, mitigation, and decision. In contrast, criminal trials adopt a simplified structure with three stages: preparation, investigation, and decision. Accordingly, court completeness is evaluated using stage-level completion (STA), action-level completion within each stage (ACT), and a binary indicator (UNI). In the criminal court setting, judgment accuracy is further assessed by evaluating predicted fines (FINE) and sentences (SEN) using normalized logarithmic deviation. Together, these fine-grained metrics provide a detailed assessment of both judgment accuracy and procedural compliance.

As shown in Tables~\ref{tab:result1} and~\ref{tab:result2}, our method demonstrates increasingly pronounced advantages in more complex court simulation scenarios. These results indicate that the proposed Explore-Verify-Memorize strategy effectively verifies intermediate reasoning steps, thereby ensuring procedural compliance and improving overall performance in judicial decision-making.

\section{Descriptions of Static Benchmarks} \label{Descriptions of Static Benchmarks}
\textbf{LexEval} adopts the Legal Cognitive Ability Taxonomy (LexCog) to systematically organize a wide range of legal tasks. The taxonomy defines six core legal abilities, including Memorization, Understanding, Logical Inference, Discrimination, Generation, and Ethics, and covers 23 tasks with a total of 14,150 questions. LexEval integrates data from existing legal benchmarks, real-world legal examination datasets, and newly annotated samples curated by legal experts, enabling a comprehensive evaluation of LLMs’ legal cognitive capabilities.

\textbf{LawBench} is designed to simulate judicial cognition along three dimensions: legal knowledge memorization, comprehension, and application. It evaluates large language models across 20 tasks, including statute recitation, legal knowledge question answering, and law article prediction. These tasks are organized into five categories: single-label classification (SLC), multi-label classification (MLC), regression, information extraction, and text generation, providing a diverse and structured assessment framework.

\textbf{UniLaw-R1-Eval} is constructed from the open-source JEC-QA dataset and proprietary data derived from the Chinese National Judicial Examination spanning 2015 to 2021, comprising 800 samples in total. The benchmark is divided into knowledge-based and case-based subsets. 

Since our focus is on knowledge-intensive legal reasoning, we select tasks 1-1, 1-2, 3-1, 3-2, and 3-4 from LexEval, 1-1, 1-2, 3-1, 3-2, and 3-6 from LawBench, and dominant domains from UniLaw-R1-Eval, including criminal procedure law, labor law, commercial law, constitutional law, and jurisprudence.

\section{Prompts in LawThinker}
We provide detailed prompts in Tables~\ref{LawThinker Instruction}, \ref{Exploration-Phase Examples}, \ref{Deep Checker Prompt}, and \ref{Checking-Phase Examples}.

\begin{figure*}[t]
\centering
\caption{LawThinker Instruction.}
\label{LawThinker Instruction}
\begin{promptbox}[title=LawThinker Instruction]
You are a legal-reasoning assistant that may call domain-specific legal tools whenever necessary.

\textbf{Available tools}\\
\verb|memory_fetch| – retrieve stored knowledge or context.\\
\verb|law_retrieval| – retrieve the top–k most relevant statutes given a natural–language query.\\
\verb|law_recommendation| – return statutes similar to the one provided.\\
\verb|charge_expansion| – expand a list of charges with related ones.\\
\verb|case_retrieval| – retrieve similar civil/criminal cases.\\
\verb|template_retrieval| – fetch a document template (e.g.\ complaint document, defence document).\\
\verb|writing_plan_generation| – generate a writing plan for the given document type.\\
\verb|procedure_retrieval| – retrieve the civil/criminal court procedure
  (\texttt{stage}=0 for the full procedure).\\

\textbf{Tool-calling format}\\
Whenever a tool is needed, output\\
\verb|<tool_call>{"name": ..., "arguments": ...}</tool_call>|.\\
The system will respond with\\
\verb|<tool_call_result> ... </tool_call_result>|.

\textbf{Example:} {examples}
\end{promptbox}
\Description{}
\end{figure*}

\begin{figure*}[t]
\centering
\caption{Exploration-Phase Examples.}
\label{Exploration-Phase Examples}
\begin{promptbox}[title=Exploration-Phase Examples]
For multi-turn QA, we recommend \texttt{memory\_fetch}, \texttt{law\_retrieval}, and
\texttt{law\_recommendation}.\\[2pt]
\verb|<tool_call>{"name":"law_retrieval",|\\
\verb|            "arguments":{"query":"Must a recusal request state reasons?", "topk":5}}</tool_call>|\\
\verb|<tool_call_result>...statute snippets...</tool_call_result>|
\verb|...|\\

For document generation, we recommend \texttt{template\_retrieval}, \texttt{plan\_generation}, and \texttt{memory\_fetch}.\\[2pt]
\verb|<tool_call>{"name":"template_retrieval",|\\
\verb|           "arguments":{"template_type":"complaint document"}}</tool_call>|\\
\verb|<tool_call_result>...template text...</tool_call_result>|\\[2pt]
\verb|<tool_call>{"name":"writing_plan_generation",|\\
\verb|           "arguments":{"document_type":"complaint document"}}</tool_call>|\\
\verb|<tool_call_result>...plan steps...</tool_call_result>|\\
\verb|<tool_call>{"name":"memory_fetch",|\\
\verb|           "arguments":{"memory_type":"knowledge"}}</tool_call>|\\
\verb|<tool_call_result>...knowledge memory...</tool_call_result>|
\verb|...|\\

For court simulation, we recommend \texttt{procedure\_retrieval}, \texttt{law\_retrieval},
\texttt{law\_recommendation}, \texttt{charge\_expansion}, and
\texttt{case\_retrieval}. Before issuing a judgment you \emph{must} gather
sufficient case facts and legal grounds.\\[2pt]
\verb|<tool_call>{"name":"procedure_retrieval",|\\
\verb|           "arguments":{"court_type":"civil court","stage":0}}</tool_call>|\\
\verb|<tool_call_result>...five stages returned...</tool_call_result>|\\[2pt]
\verb|<tool_call>{"name":"law_retrieval",|\\
\verb|           "arguments":{"query":"case summary", "topk":5}}</tool_call>|\\
\verb|<tool_call_result>...candidate statutes...</tool_call_result>|\\[2pt]
\verb|<tool_call>{"name":"law_recommendation",|\\
\verb|           "arguments":{"law":"Criminal Law Art.201"}}</tool_call>|\\
\verb|<tool_call_result>...related articles...</tool_call_result>|\\[2pt]
\verb|<tool_call>{"name":"charge_expansion",|\\
\verb|           "arguments":{"charges":["charge"]}}</tool_call>|\\
\verb|<tool_call_result>...alternative charges...</tool_call_result>|\\[2pt]
\verb|<tool_call>{"name":"case_retrieval",|\\
\verb|           "arguments":{"type":"civil","query":"key facts"}}</tool_call>|\\
\verb|<tool_call_result>...precedents...</tool_call_result>|
\verb|...|\\
\end{promptbox}
\Description{}
\end{figure*}

\begin{figure*}[t]
\centering
\caption{DeepVerifier Prompt.}
\label{Deep Checker Prompt}
\begin{promptbox}[title=DeepVerifier Prompt]
You are a deep-analysis legal assistant.  
Given (i) the user’s last query or response, (ii) the current reasoning trace, and
(iii) the result of the exploration step, decide whether the retrieved
information is correct and relevant; decide whether more exploration is needed;
and store any key facts or statutes.

\textbf{Tool principles}\\
1. Choose the appropriate \emph{checking} tool to verify the accuracy and
relevance of retrieved knowledge.\\
2. Summarize key information and store it with \texttt{memory\_store}.\\

\textbf{Available tools}\\
\verb|memory_fetch| \quad retrieve stored knowledge / context\\
\verb|memory_store| \quad store key knowledge / context\\
\verb|law_article_check| \quad verify law article content\\
\verb|fact_law_relevance_check| \quad check law applicability to facts\\
\verb|charge_law_consistency_check| \quad check charge-law consistency\\
\verb|search_query_rewrite| \quad rewrite law retrieval queries\\
\verb|document_format_check| \quad check document format\\
\verb|procedure_check| \quad Check procedural compliance\\

\textbf{Tool-calling format}\\
\verb|<tool_call>{"name": ..., "arguments": ...}</tool_call>|\\
\verb|<tool_call_result> ... </tool_call_result>|.\\

\textbf{Example:} {examples}
\end{promptbox}
\Description{}
\end{figure*}

\begin{figure*}[t]
\centering
\caption{Checking-Phase Examples.}
\label{Checking-Phase Examples}
\begin{promptbox}[title=Checking-Phase Examples]
\textbf{Multi-turn QA}\\
\verb|<tool_call>{"name":"fact_law_relevance_check",|\\ 
\verb|           "arguments":{"fact":"...","law":"Art.201"}}</tool_call>|\\
\verb|<tool_call_result> ... </tool_call_result>|.\\
\verb|<tool_call>{"name":"law_query_rewrite",|\\
\verb|           "arguments":{"query":"original","context":"case background"}}</tool_call>|\\
\verb|<tool_call_result> ... </tool_call_result>|.\\

\textbf{Document Generation}\\
\verb|<tool_call>{"name":"document_format_check",|\\
\verb|           "arguments":{"document_type":"complaint document","document":"<draft text>"}}|\\ 
\verb|</tool_call>|\\
\verb|<tool_call_result> ... </tool_call_result>|.\\

\textbf{Court Simulation}\\
\verb|<tool_call>{"name":"procedure_check",|\\
\verb|           "arguments":{"court_type":"civil court"}}</tool_call>|\\
\verb|<tool_call_result> ... </tool_call_result>|.\\
\verb|<tool_call>{"name":"law_check","arguments":{"law_name":"Art.201"}}</tool_call>|\\
\verb|<tool_call_result> ... </tool_call_result>|.\\

\end{promptbox}
\Description{}
\end{figure*}

\end{document}